\documentclass[sigconf,screen]{acmart}

\usepackage{multirow}
\usepackage{tabularx}

\DeclareMathOperator*{\concat}{\scalebox{1}[1.5]{$\parallel$}}

\setcopyright{acmcopyright}
\copyrightyear{2020}
\acmYear{2020}
\acmDOI{10.1145/3394171.3413600}

\acmConference[MM '20]{Proceedings of the 28th ACM International Conference on Multimedia}{October 12--16, 2020}{Seattle, WA, USA}
\acmBooktitle{Proceedings of the 28th ACM International Conference on Multimedia (MM '20), October 12--16, 2020, Seattle, WA, USA}
\acmPrice{15.00}
\acmISBN{978-1-4503-7988-5/20/10}

\begin{document}

% camera ready version
% \fancyhead{}

\title{ConsNet: Learning Consistency Graph for Zero-Shot Human-Object Interaction Detection}

\author{Ye Liu$^{1}$, Junsong Yuan$^{2}$, Chang Wen Chen$^{2,3,4}$}
\affiliation{
\institution{$^1$Department of Resource and Environmental Sciences, Wuhan University, China}
\institution{$^2$Department of Computer Science and Engineering, State University of New York at Buffalo, USA}
\institution{$^3$Peng Cheng Laboratory, China}
\institution{$^4$School of Science and Engineering, The Chinese University of Hong Kong, Shenzhen, China}}
\affiliation{ye-liu@whu.edu.cn, \{\href{mailto:jsyuan@buffalo.edu}{\textcolor{black}{jsyuan}},\href{mailto:chencw@buffalo.edu}{\textcolor{black}{chencw}}\}@buffalo.edu}

\renewcommand{\authors}{Ye Liu, Junsong Yuan, and Chang Wen Chen}
\renewcommand{\shortauthors}{\authors}

\begin{abstract}

We consider the problem of Human-Object Interaction (HOI) Detection, which aims to locate and recognize HOI instances in the form of $\langle human, action, object \rangle$ in images. Most existing works treat HOIs as individual interaction categories, thus can not handle the problem of long-tail distribution and polysemy of action labels. We argue that multi-level consistencies among objects, actions and interactions are strong cues for generating semantic representations of rare or previously unseen HOIs. Leveraging the compositional and relational peculiarities of HOI labels, we propose \textit{ConsNet}, a knowledge-aware framework that explicitly encodes the relations among objects, actions and interactions into an undirected graph called \textit{consistency graph}, and exploits Graph Attention Networks (GATs) to propagate knowledge among HOI categories as well as their constituents. Our model takes visual features of candidate human-object pairs and word embeddings of HOI labels as inputs, maps them into visual-semantic joint embedding space and obtains detection results by measuring their similarities. We extensively evaluate our model on the challenging V-COCO and HICO-DET datasets, and results validate that our approach outperforms state-of-the-arts under both fully-supervised and zero-shot settings. Code is available at \href{https://github.com/yeliudev/ConsNet}{https://github.com/yeliudev/ConsNet}.

\end{abstract}




\maketitle

\section{Introduction}

Beyond detecting individual human or object instances in images, it is crucial for machines to also recognize how they interact with each other, which can be essential cues to understand the human-centric visual world. The task of Human-Object Interaction (HOI) Detection aims to locate and recognize HOI instances in images. For example, detecting $\langle human, feed, cat \rangle$ refers to locating ``human'' and ``cat'', as well as predicting the action ``feed'' for this human-object pair. Instead of inferring ambiguous spatial relations among objects, e.g. ``cat is on the bed'', HOI detection plays a pivotal role to understand \textit{what is happening} in the scene. Studying HOIs can benefit many down-stream visual understanding tasks including image captioning \cite{li2017scene}, image retrieval \cite{xu2017scene}, and visual question answering \cite{goyal2017making}.

\begin{figure}
\centering
\includegraphics[width=\linewidth]{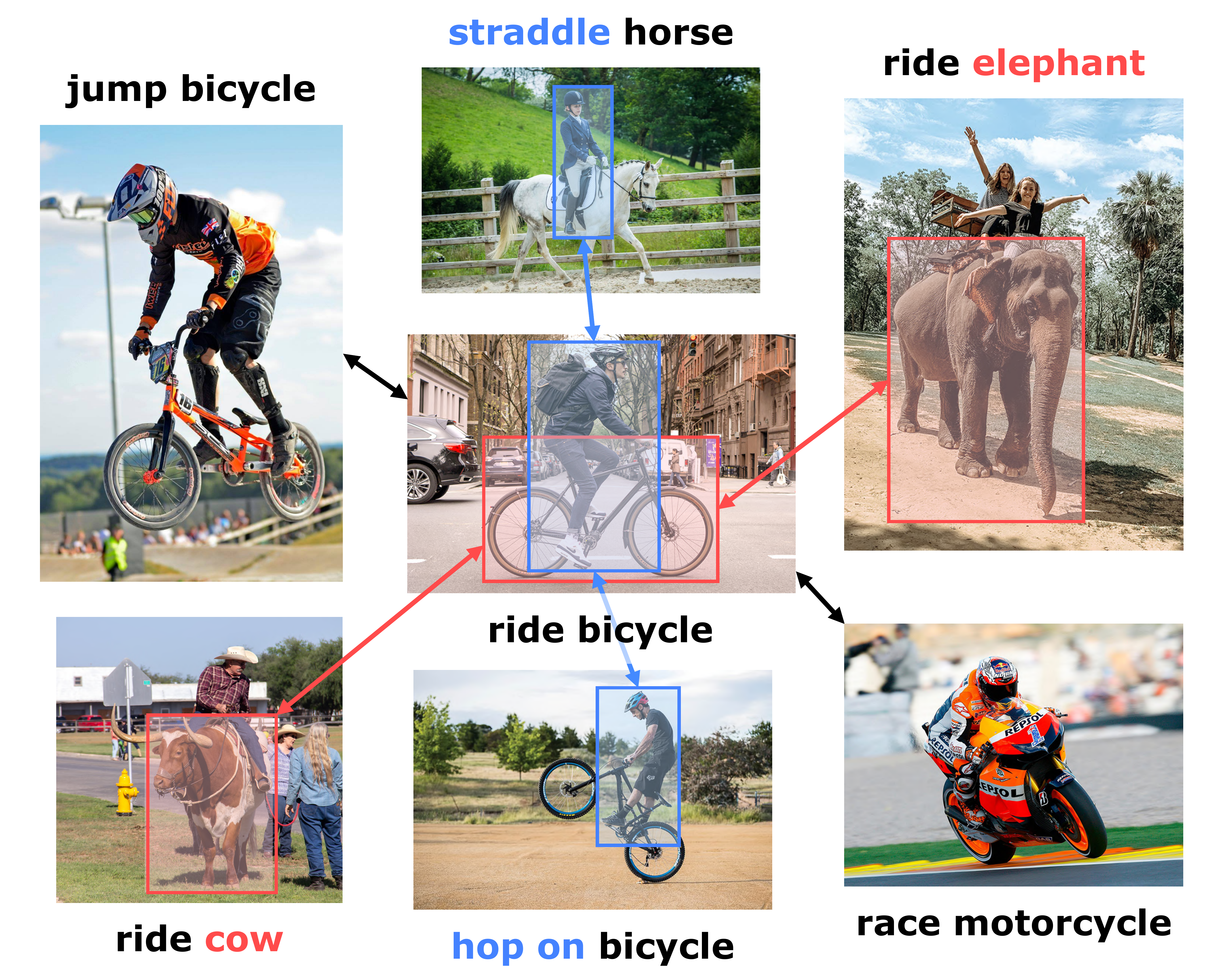}
\caption{Illustration of knowledge-aware human-object interaction detection. Red, blue and black lines represent functionally similar objects, behaviorally similar actions, and holistically similar interactions. We argue that successful detection of an HOI should benefit from the knowledge obtained from similar objects, actions, and interactions.}
\Description{Illustration of knowledge-aware human-object interaction detection.}
\label{fig1}
\end{figure}

\begin{figure}
\centering
\includegraphics[width=\linewidth]{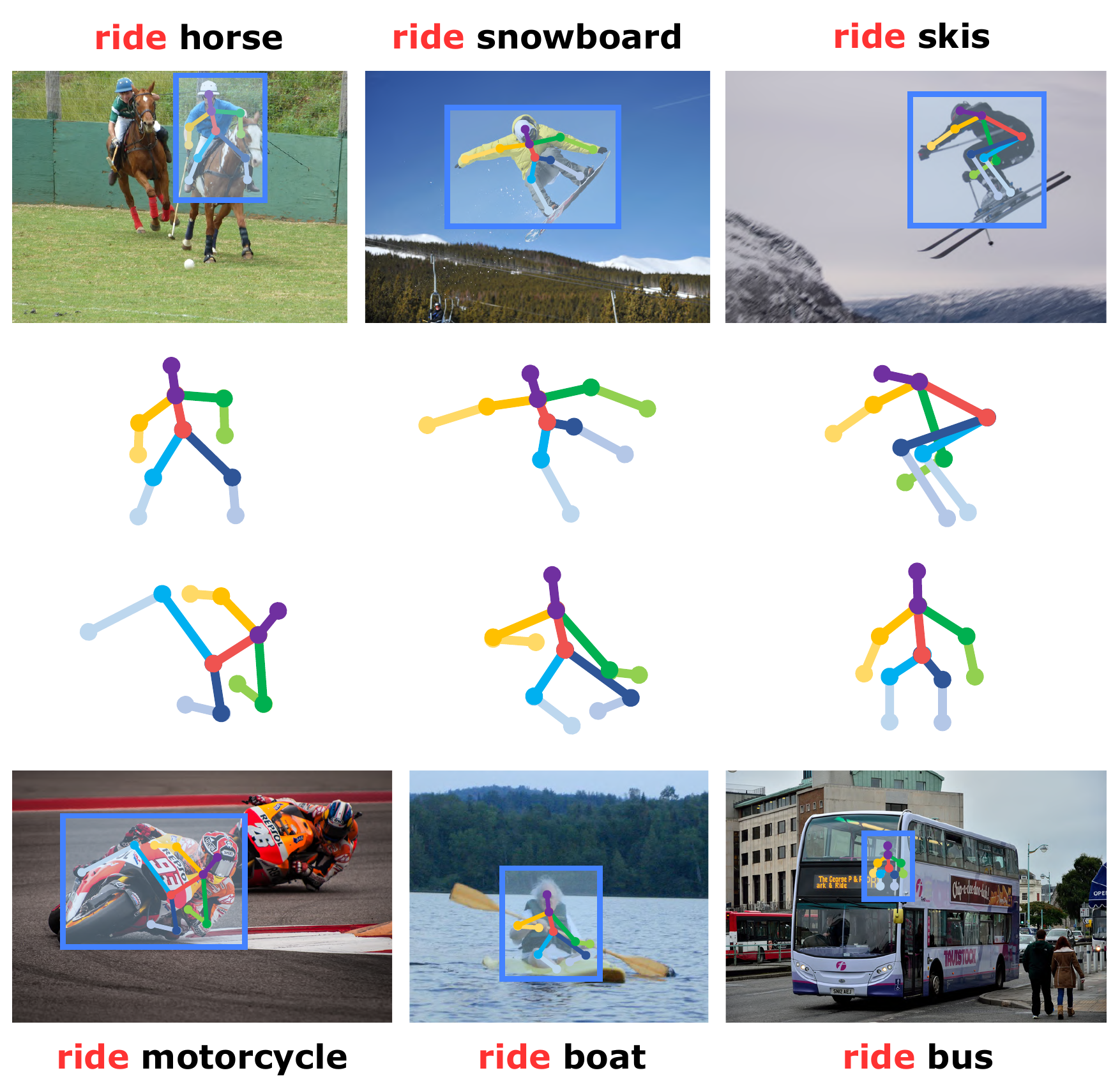}
\caption{Polysemy of action labels. All the HOIs above share the same action label ``ride'', but the actual implications of these actions are inconsistent, as can be seen from the inferred human poses.}
\Description{Polysemy of action labels.}
\label{fig2}
\end{figure}

Most existing works on HOI detection \cite{gkioxari2018detecting,shen2018scaling,gao2018ican,li2019transferable,wang2019deep,wan2019pose,gupta2019no} treat HOIs as individual interaction categories and focus on mining visual representations of human-object pairs to improve classification performances. Despite previous successes, these conventional approaches still face two challenges. First, compared with other action-based recognition tasks, what makes HOI detection challenging is that labels of HOIs are fine-grained and are related to the specific object category. The quadratic number of combinations of actions and objects brings prohibitive annotation cost. Hence, non-compositional methods \cite{chao2018learning,gao2018ican,li2019transferable,qi2018learning,wang2019deep,wan2019pose} are largely restricted by the coverage and long-tail distribution of exhaustive HOI annotations. Second, the compositional peculiarity of HOI labels also leads to the polysemy of action labels. As an example shown in Figure~\ref{fig2}, collocated with different objects, the actual implications of action ``ride'' are sometimes inconsistent. Such phenomenon brings ambiguities and extra challenges to compositional methods \cite{gkioxari2018detecting,shen2018scaling,gupta2019no,bansal2019detecting}.

In this work, we address the above two challenges by proposing a knowledge-aware approach (as shown in Figure~\ref{fig1}) for HOI detection. For the first challenge, we claim that the key to dealing with the imbalance and scarcity of HOI training samples is to distill knowledge obtained from non-rare categories, and transfer it to rare or unseen ones. Considering that humans have the ability to perceive unseen interactions, e.g. $\langle human, ride, elephant \rangle$, because they can make use of their common sense to \textit{imagine} what it would be like based on similar HOIs such as $\langle human, ride, bicycle \rangle$ and $\langle human, feed, elephant \rangle$, as well as similar actions or objects such as ``sit on'' or ``horse''. To jointly capture the compositional peculiarities and multi-level similarities among HOIs, we define three types of consistencies at different granularities. At unigram level, we introduce \textbf{functional consistency} which depicts the functional similarities among objects, and \textbf{behavioral consistency} that represents the similarities of human behaviors when performing different actions. At trigram level, we present \textbf{interactional consistency}, which denotes the holistic similarities among HOIs. We further construct an undirected graph, namely \textbf{consistency graph}, to explicitly encode these relations. Each node in the consistency graph represents an HOI label or one of its entities. The three types of consistencies are encoded as edges among the nodes. That is, two object, action or interaction nodes are linked if they have whichever the consistencies above. We then use word embeddings of HOI labels as input features of nodes, and exploit recently introduced Graph Attention Networks (GATs) \cite{velickovic2018graph} to perform message passing on the consistency graph, enabling the model to learn semantic representations of HOIs in a transductive manner.

When it comes to the second challenge, we argue that an appropriate perception of HOI should benefit from both unigram and trigram representations. Take the HOI $\langle human, ride, bicycle \rangle$ for instance. At unigram level, we ought to make sure that the subject is a human, the object is a bicycle, and the subject is performing the action ``ride''. At trigram level, we should also deem that the human-object pair is performing the right interaction holistically. In our model, HOI detection scores are estimated based on the similarities between visual and semantic embeddings of human, object, action, and interaction. Such a decomposition strategy helps capture implications of HOIs at multiple granularities, thus can better handle the polysemy of action labels. Moreover, our model has the ability to transfer knowledge from familiar HOIs to HOIs with unseen actions, objects, or action-object combinations. Note that detecting HOIs with unseen actions may not be performed by previous methods.

The main contributions of our work are as follows:

\begin{itemize}
\item We propose a knowledge-aware approach to model relations among HOIs at both unigram and trigram level, and exploit Graph Attention Networks to predict semantic representations of HOIs based on their word embeddings.
\item We introduce a data-driven method to estimate consistencies and construct the consistency graph using visual-semantic representations of HOI labels, which can jointly capture visual and semantic features of HOIs.
\item Our approach outperforms state-of-the-arts under both fully-supervised and zero-shot settings on the challenging V-COCO and HICO-DET datasets. Further experiments also show that our model has the ability to detect HOIs with unseen actions, which may not be performed by previous methods.
\end{itemize}

\section{Related Works}

\begin{figure*}
\centering
\includegraphics[width=\textwidth]{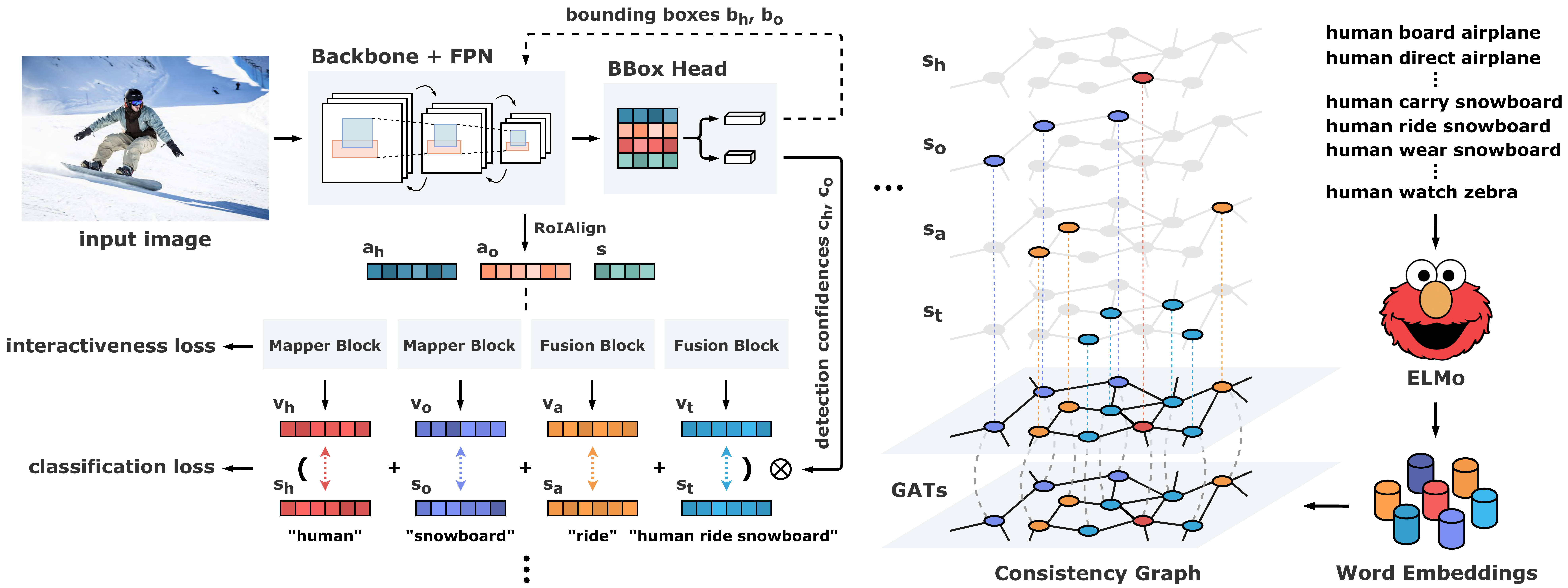}
\caption{Overall architecture of our framework. The input image is fed into a pre-trained object detector to obtain bounding boxes $b_h$, $b_o$ with detection confidences $c_h$, $c_o$ of humans and objects. The bounding boxes are then used to crop visual features $a_h$, $a_o$ from FPN and compute spatial configuration $s$. Subsequently, visual embedding network maps these features into multi-level visual embeddings $v_h, v_o, v_a, v_t$. On the other side, semantic embedding network encodes HOI labels into vectors using a pre-trained language model. The word embeddings serve as input features of nodes in the consistency graph. By performing GATs, these features are propagated among neighboring nodes and be transformed into semantic embeddings $s_h, s_o, s_a, s_t$. The HOI detection results are then generated by measuring the similarities among visual embeddings and semantic embeddings.}
\Description{Overall architecture of our framework.}
\label{fig3}
\end{figure*}

\paragraph{Human-Object Interaction Detection}

Human-Object Interaction Detection plays a crucial role in human-centric scene understanding since the problem was first introduced by Gupta and Malik \cite{gupta2015visual}. Most previous works can be divided into compositional methods \cite{gkioxari2018detecting,shen2018scaling,gupta2019no,bansal2019detecting} and non-compositional methods \cite{chao2018learning,gao2018ican,li2019transferable,qi2018learning,wang2019deep,wan2019pose}. Compositional methods learn separate detectors for objects and actions, then fuse the confidences to generate HOI detection results. However, these approaches suffer from the polysemy of action labels. Non-compositional methods avoid this problem by predicting fine-grained HOI labels directly, but they are restricted by the long-tail distribution of HOI categories. Recently introduced hybrid model \cite{peyre2019detecting} has shown that using multi-granularity representations of HOIs may solve the above contradiction. Nonetheless, all these methods ignore the implicit relations among HOIs, thus we extend the hybrid model by incorporating common sense knowledge for generating semantic embeddings.

\paragraph{Graph Neural Networks}

The past few years have witnessed the rapid development of representation learning on graphs \cite{jie2018graph}. The majority of these methods are under the Message Passing Neural Networks (MPNN) framework \cite{gilmer2017neural} which decomposes the pipeline into message functions, vertex update functions, and readout functions. Kipf \textit{et al.} \cite{kipf2016semi} extend the convolution operation \cite{lecun1998gradient} from euclidean data to non-euclidean data and proposed Graph Convolutional Networks (GCNs). Wu \textit{et al.} \cite{wu2019simplifying} introduced SGCs to simplify GCNs by removing the non-linearities and merging the weights. Hamilton \textit{et al.} \cite{hamilton2017inductive} proposed GraphSAGE to realize inductive learning on graphs. In this work, we exploit Graph Attention Networks (GATs) \cite{velickovic2018graph} that incorporate multi-head attention mechanism to model the relations of neighboring nodes. The learned attention coefficients in GATs serve as the weights of consistencies.

\paragraph{Zero-Shot Learning}

Most recent zero-shot learning methods can be divided into two protocols \cite{wang2019survey}. One is to learn semantic representations of categories that can be mapped to visual classifiers \cite{changpinyo2016synthesized,changpinyo2017predicting}. The other is to make use of knowledge graphs to distill the knowledge \cite{chen2013neil,deng2014large}. In this work, with the help of GNNs and language models, we learn the explicit and implicit knowledge of HOIs from consistency graph and word embeddings respectively.

\section{Approach}

In this section, we introduce our approach on knowledge-aware HOI detection. As illustrated in Figure~\ref{fig3}, the entire framework can be divided into two sub-modules, namely visual embedding network and semantic embedding network. These sub-modules map visual representations of human-object pairs and word embeddings of HOI labels into visual-semantic joint embedding space. HOI detection results are then generated by measuring similarities between visual and semantic embeddings.

\begin{figure}
\centering
\includegraphics[width=0.95\linewidth]{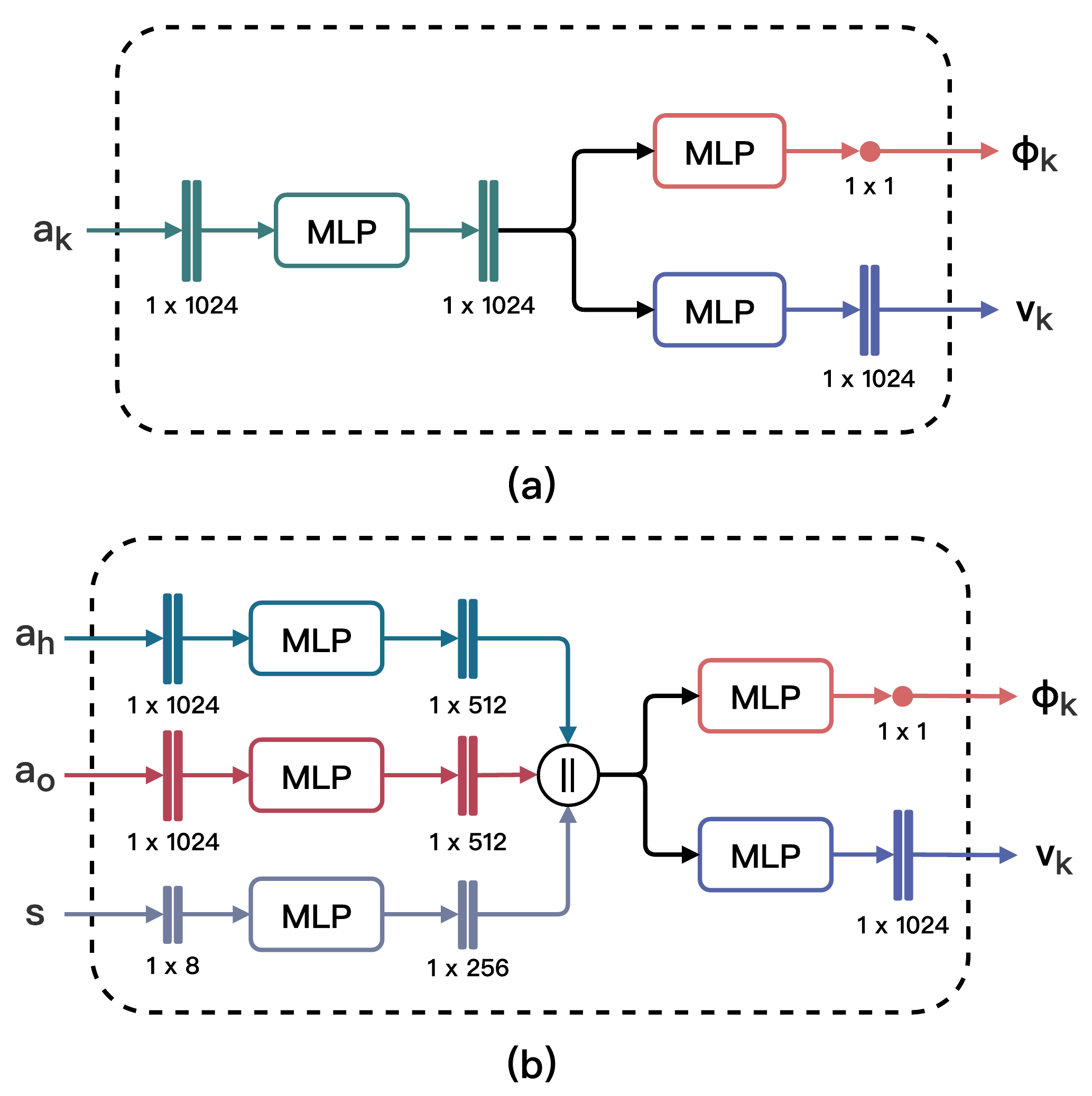}
\caption{Detailed architecture of visual embedding network. a) Mapper block takes visual features of human or object $a_k, k \in \{h, o\}$ as inputs, and predicts visual embeddings $v_k, k \in \{h, o\}$ as well as interactiveness $\varphi_k, k \in \{h, o\}$. b) Fusion block takes human features $a_h$, object features $a_o$ and their spatial configuration $s$ as inputs, and estimates visual embeddings of action or interaction $v_k, k \in \{a, t\}$, together with interactiveness $\varphi_k$.}
\Description{Detailed architecture of visual embedding network.}
\label{fig4}
\end{figure}

\subsection{Overview}

Given an image $x$ and a set of HOI categories of interest $\mathcal{H} = \{1,...,C\}$, the task of human-object interaction detection is to detect all the human-object pairs in $x$, where the humans and objects are participating one or multiple pre-defined interactions. The outputs of HOI detection would be a set of tuples $\mathcal{T} = \{\langle b_h, b_o, y_{h,o} \rangle\}$, where $b_h, b_o \in \mathbb{R}^4$ denotes bounding boxes of the human and the object, and $y_{h, o}$ represents a vector where $y^i_{h, o} \in \{0,1\}$ indicates whether the HOI class $i$ is assigned to this human-object pair. Note that a person may have several interactions with multiple objects simultaneously, thus different HOIs may share the same human, action or object.

We adopt a three-stage HOI detection pipeline by generating a set of human-object pairs as candidates, filtering out non-interactive ones and classifying the remainders into multiple interaction categories. In the first stage, a pre-trained object detector is used to collect bounding boxes of humans $B_h$ and objects $B_o$, along with their corresponding detection confidences $C_h$, $C_o$. We only keep top $N_k$ detections with confidences $c_k$ higher than a threshold $\theta_k$, where $k \in \{h, o\}$ denotes human or object. The candidates are then obtained by pairing up all the humans and objects extensively.

Recent works have shown that in most cases, the majority of humans and objects in an image are not interacting with each other. Such a severe imbalance between positive and negative candidates makes HOI classification challenging. To address this problem, Li \textit{et al.} \cite{li2019transferable} proposed the strategy of non-interactive suppression (NIS) to filter out and suppress potential non-interactive candidates. In the second stage, we predict the class-irrelevant interactiveness $\varphi_{h,o}$ for each candidate by
\begin{gather}
\varphi_{h,o} = \sigma(\sum_{k \in \{h, o, a, t\}} \varphi_k)
\end{gather}
where $\sigma(\cdot)$ denotes the Sigmoid function and $\varphi_k, k \in \{h, o, a, t\}$ indicates the interactiveness score at human, object, action or interaction level. Candidates with interactiveness $\varphi_{h,o}$ lower than a threshold $\theta_{h,o}$ would be discarded. The remaining ones are then fed into HOI classifier for further interaction classification.

In the third stage, we classify the candidates into HOI categories in a knowledge-aware manner. For each candidate, the confidence of assigning HOI class $i$ to it can be given by
\begin{gather}
P(y_i = 1 | x, b_h, b_o, c_h, c_o) = r^i_{h, o} \cdot \varphi_{h, o} \cdot c_h \cdot c_o
\end{gather}
where $r^i_{h, o}$ is the HOI classification score given by the HOI classifier. Interactiveness $\varphi_{h, o}$, human detection confidence $c_h \in C_h$ and object detection confidence $c_o \in C_o$ serve as suppression terms on potential non-interactive or non-existent candidates. The HOI classification score $r^i_{h, o}$ can be given by
\begin{gather}
r^i_{h, o} = \sigma(\sum_{k \in \{h, o, a, t\}} \frac{v_k \cdot s^i_k}{\|v_k\|_2 \cdot \|s^i_k\|_2} \cdot \gamma)
\end{gather}
where $v_k$ denotes visual embeddings of the candidate, including human $v_h$, object $v_o$, action $v_a$, and interaction $v_t$. $s_k^i$ represents semantic embeddings of these entities for HOI class $i$. We treat $s_k$ as templates of HOIs and measure the distance among visual and semantic embeddings by computing cosine similarities. Note that we also add a scale factor $\gamma$ to control the range of outputs.

The visual embeddings $v_k$, interactiveness $\varphi_{h, o}$ and semantic embeddings $s_k$ are generated by visual embedding network and semantic embedding network. Details of the embedding networks are explained in the following sections.

\begin{figure*}
\centering
\includegraphics[width=0.95\textwidth]{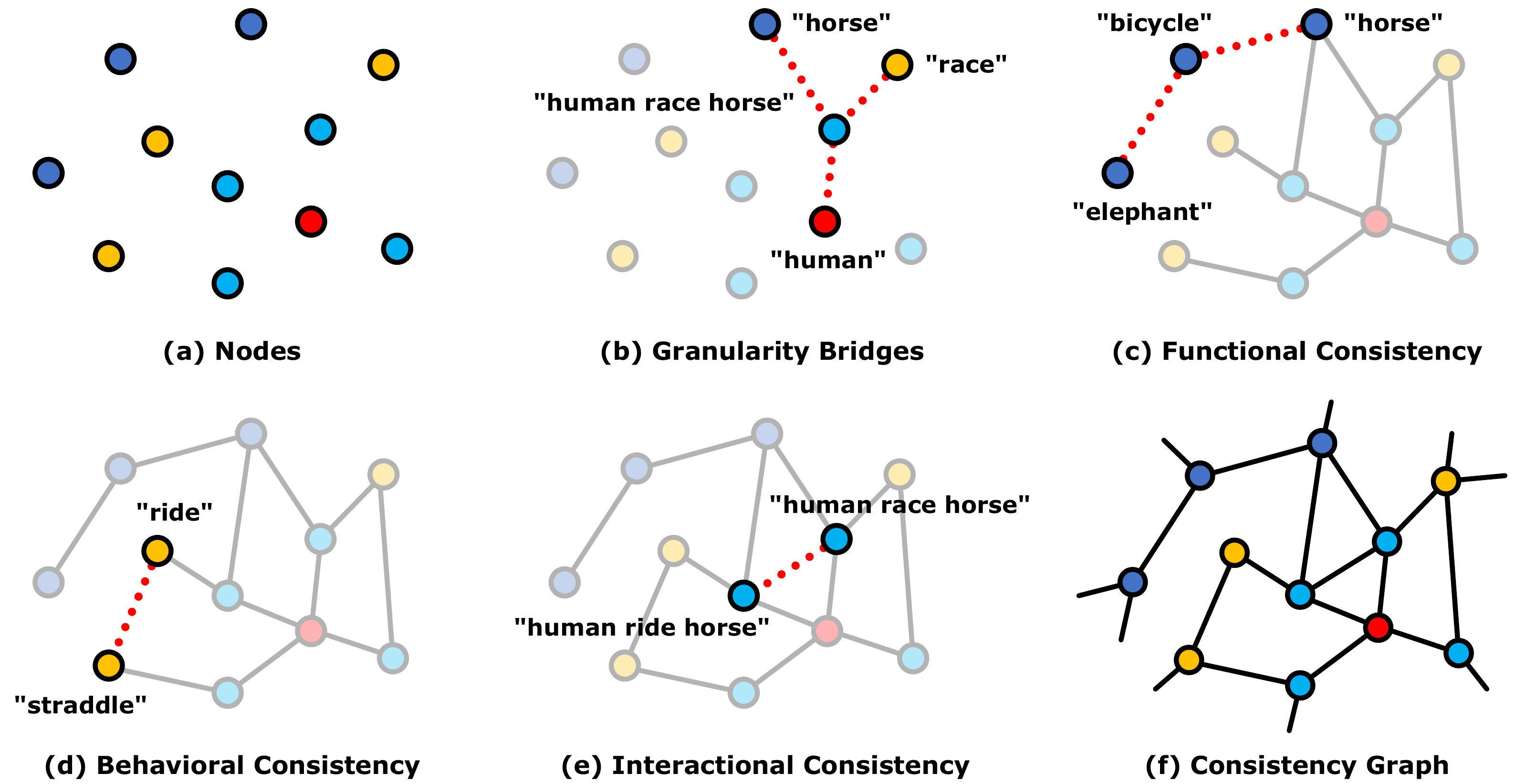}
\caption{Pipeline of constructing the consistency graph. a) Consistency graph contains human, object, action, and interaction nodes. b) Each interaction node is linked with its entity nodes. c) Functional consistencies are represented by object-object connections. d) Behavioral consistencies are represented by action-action connections. e) Interactional consistencies are represented by interaction-interaction connections. f) Generalize the rules above and build consistency graph.}
\Description{Pipeline of constructing the consistency graph.}
\label{fig5}
\end{figure*}

\subsection{Visual Embedding Network}\label{3.2}

Visual embedding network takes image $x$ as well as bounding boxes of human and object $b_h$, $b_o$ as inputs, and generates visual embeddings of human $v_h$, object $v_o$, action $v_a$, and interaction $v_t$. These visual embeddings are constructed based on visual features of human $a_h$, object $a_o$, and their spatial configuration $s$. We adopt ResNet-50-FPN \cite{he2016deep,lin2017feature}, which can be shared with the object detector, as the feature extractor. We obtain the visual features of human and object by cropping the appropriate level of feature map from FPN using RoIAlign \cite{he2017mask} according to their bounding boxes. Spatial configuration of a candidate is computed by
\begin{gather}
s = \concat_{k \in \{h, o\}} (\frac{x^k_1 - d_x}{\psi} \| \frac{x^k_2 - d_x}{\psi} \| \frac{y^k_1 - d_y}{\psi} \| \frac{y^k_2 - d_y}{\psi})
\end{gather}
where $\|$ denotes concatenation operation, $x^k_1, x^k_2, y^k_1, y^k_2, k \in \{h, o\}$ are coordinates of the human or object bounding box, $(d_x, d_y)$ and $\psi$ represent the origin and area of the union box respectively. The computed spatial configuration $s$ would be an $1 \times 8$ vector. We hypothesize that visual embeddings of human and object can be predicted by their own visual features $a_k, k \in \{h, o\}$, while visual embeddings of action and interaction are jointly affected by visual features of human and object $a_k, k \in \{h, o\}$ as well as their spatial configuration $s$.
\begin{gather}
P(\varphi_m, v_m\ |\ x, b_h, b_o) = P(\varphi_m, v_m\ |\ a_m), m \in \{h, o\}\\
P(\varphi_n, v_n\ |\ x, b_h, b_o) = P(\varphi_n, v_n\ |\ a_h, a_o, s), n \in \{a, t\}
\end{gather}
Based on the hypotheses above, we introduce two types of embedding blocks, i.e. mapper block and fusion block, to predict interactiveness $\varphi_{h, o}$ and generate visual embeddings $v_k, k \in \{h, o, a, t\}$ for candidates. Details of the embedding blocks are described in section \ref{3.2.1} and \ref{3.2.2}.

\subsubsection{Mapper Block}\label{3.2.1}

As shown in Figure~\ref{fig4} (a), mapper block only takes visual features of the human or object as inputs. These visual features are first transformed into hidden states by a multi-layer perceptron (MLP). After that, two MLPs are used to map the dimensions of hidden states to $1 \times 1$ and $1 \times 1024$ respectively. The two outputs are interactiveness $\varphi_k$ and visual embeddings $v_k$.

\subsubsection{Fusion Block}\label{3.2.2}

As described in Figure~\ref{fig4} (b), fusion block receives visual features of the human $a_h$, object $a_o$ and their spatial configuration $s$ as inputs, and does the same job as mapper blocks. The only difference is that dimensions of $a_h$, $a_o$ and $s$ are mapped to $1 \times 512$, $1 \times 512$ and $1 \times 256$ respectively using MLPs in advance. The concatenation of the mapped features serves as joint features of the human-object pair and be used to estimate $\varphi_k$ and $v_k$.

\subsection{Semantic Embedding Network}

To jointly capture multi-level consistencies among HOIs, we incorporate a knowledge graph, namely \textit{consistency graph}, into the semantic embedding network to help generate semantic embeddings of HOI categories.

\subsubsection{Constructing the Graph}

Instead of using a large-scale knowledge graph, we distill the knowledge and construct a much smaller one, which only contains consistencies and compositional relations among HOIs and their entities. As illustrated in Figure~\ref{fig5}, each HOI category refers to three entity nodes and one interaction node in the consistency graph. HOIs with shared entities would share the entity nodes as well. For instance, $\langle human, ride, bicycle \rangle$ and $\langle human, ride, horse \rangle$ are represented by four entity nodes ``human'', ``ride'', ``bicycle'', and ``horse'', as well as two interaction nodes ``human ride bicycle'' and ``human ride horse''.

\begin{table}
\caption{Role Detection results on V-COCO dataset under fully-supervised settings.}
\label{tab:1}
\begin{tabularx}{0.82\linewidth}{p{2.4cm}<{\raggedright}|p{2.4cm}<{\raggedright}|p{1cm}<{\centering}}
\toprule
\textbf{Method}&\textbf{Backbone}&\boldmath{$\rm{mAP_{role}}$}\\
\midrule
Gupta \textit{et al.} \cite{gupta2015visual}&ResNet-50-FPN&31.8\\
InteractNet \cite{gkioxari2018detecting}&ResNet-50-FPN&40.0\\
GPNN \cite{qi2018learning}&DCN&44.0\\
iCAN \cite{gao2018ican}&ResNet-50&45.3\\
TIN-${\rm{RP_{T2}C_D}}$ \cite{li2019transferable}&ResNet-50&48.7\\
BAR-CNN \cite{kolesnikov2019detecting}&Inception-ResNet&43.6\\
Wang \textit{et al.} \cite{wang2019deep}&ResNet-50&47.3\\
PMFNet \cite{wan2019pose}&ResNet-50&52.0\\
IP-Net \cite{wang2020learning}&Hourglass-104&51.0\\
VSGNet \cite{ulutan2020vsgnet}&ResNet-152&51.8\\
\textbf{ConsNet} (ours)&ResNet-50-FPN&\textbf{53.2}\\
\bottomrule
\end{tabularx}
\end{table}

We first add edges among interaction nodes and their corresponding entity nodes, which serve as bridges among different levels of consistencies. The other edges are defined based on the consistencies among objects, actions, and interactions. That is, if two nodes are semantically consistent with each other, an edge would be added to enable message passing between them. We estimate the multi-level consistencies using cosine similarity by
\begin{gather}
\Theta_k(i, j) = \frac{z^i_k \cdot z^j_k}{\|z^i_k\|_2 \cdot \|z^j_k\|_2}, k \in \{a, o, t\}
\end{gather}
where $k \in \{a, o, t\}$ indicates the type of the node, $\Theta_k(i, j)$ denotes the consistency between node $i$ and $j$, $z^i_k$ and $z^j_k$ represent visual-semantic joint features of the two nodes respectively. For each node, we link itself with only top $\varepsilon_k$ consistent nodes.

\begin{table}
\renewcommand\tabcolsep{4pt}
\caption{HOI Detection results on HICO-DET dataset under fully-supervised settings. \textit{R} and \textit{H} represent ResNet and Hourglass respectively.}
\label{tab:2}
\begin{tabularx}{\linewidth}{p{2.4cm}<{\raggedright}|p{1.45cm}<{\raggedright}|p{1cm}<{\centering}p{1cm}<{\centering}p{1cm}<{\centering}p{1cm}<{\centering}}
\toprule
\textbf{Method}&\textbf{Backbone}&\textbf{Full}&\textbf{Rare}&\multicolumn{2}{l}{\hspace{-0.15cm}\textbf{Non-Rare}}\\
\midrule
Shen \textit{et al.} \cite{shen2018scaling}&VGG-19&6.46&4.24&7.12\\
HO-RCNN \cite{chao2018learning}&CaffeNet&7.81&5.37&8.54\\
InteractNet \cite{gkioxari2018detecting}&R-50-FPN&9.94&7.16&10.77\\
GPNN \cite{qi2018learning}&DCN&13.11&9.34&14.23\\
iCAN \cite{gao2018ican}&R-50&14.84&10.45&16.15\\
TIN-${\rm{RP_{T2}C_D}}$ \cite{li2019transferable}&R-50&17.22&13.51&18.32\\
HOID \cite{wang2020discovering}&R-50-FPN&17.85&12.85&19.34\\
Wang \textit{et al.} \cite{wang2019deep}&R-50-FPN&16.24&11.16&17.75\\
Gupta \textit{et al.} \cite{gupta2019no}&R-152&17.18&12.17&18.68\\
PMFNet \cite{wan2019pose}&R-50-FPN&17.46&15.65&18.00\\
Peyre \textit{et al.} \cite{peyre2019detecting}&R-50-FPN&19.40&15.40&20.75\\
IP-Net \cite{wang2020learning}&H-104&19.56&12.79&21.58\\
VSGNet \cite{ulutan2020vsgnet}&R-152&19.80&16.05&20.91\\
\textbf{ConsNet} (ours)&R-50-FPN&\textbf{22.15}&\textbf{17.55}&\textbf{23.52}\\
\midrule
Bansal \textit{et al.} \cite{bansal2019detecting}&R-101&21.96&16.43&23.62\\
PPDM \cite{liao2020ppdm}&H-104&21.73&13.78&24.10\\
\textbf{ConsNet-F} (ours)&R-50-FPN&\textbf{25.94}&\textbf{19.35}&\textbf{27.91}\\
\bottomrule
\end{tabularx}
\end{table}

We propose a data-driven approach to generate the joint features of nodes. First, we collect all the visual features of humans and objects in the dataset using a pre-trained object detector. These features are regarded as visual representations of actions and objects respectively. We then compute the average of all the visual representations with the same label to obtain the universal visual representations of these categories. Second, we adopt a pre-trained language model to generate word embeddings of node labels. Note that a label may contain multiple words, we fuse the word embeddings by computing their weighted sum. After collecting universal visual representations and word embeddings of node labels, we obtain the joint features of nodes by
\begin{gather}
z_k = (\rho_v \cdot \frac{q_k}{\|q_k\|_2})\ \|\ (\rho_s \cdot \frac{e_k}{\|e_k\|_2}), k \in \{a, o, t\}
\end{gather}
where $q_k$ and $e_k$ are visual and semantic representations of node labels, $\rho_v$ and $\rho_s$ are the weights of the representations. The L-2 normalized, re-weighted and concatenated visual-semantic representations are then used to estimate multi-level consistencies.

\subsubsection{Learning to Aggregate Semantic Representations}

Graph Attention Networks (GATs) \cite{velickovic2018graph} are first introduced for the task of semi-supervised node classification. Instead of simply averaging the features of neighboring nodes like GCNs \cite{kipf2016semi} or SGCs \cite{wu2019simplifying}, GATs aggregate node features using a self-attention strategy. A single-level GAT layer can be represented as
\begin{gather}
h_i = \concat^D_{d = 1} \tau(\sum_{j \in N_i} \mu^d_{i, j} \cdot \mathcal{W}^d \cdot h_j)
\end{gather}
where $h_i$ and $h_j$ denote the hidden states of node $i$ and $j$, $D$ indicates the number of attention heads, $\tau$ is the ReLU nonlinearity, $N_i$ represents the collection of node $i$ and its neighbours, $\mu^d_{i, j}$ is the attention coefficient learned by the model and $\mathcal{W}^d$ refers to the weights of this layer. In order to fix the output dimensions of the last GAT layer, we replace its concatenation with average operation. The attention coefficient $\mu^d_{i, j}$ can be predicted by
\begin{gather}
\mu^d_{i, j} = \frac{exp(\Gamma(\mathcal{W}^d \cdot h_i\ \|\ \mathcal{W}^d \cdot h_j))}{\sum_{k \in N_i} exp(\Gamma(\mathcal{W}^d \cdot h_i\ \|\ \mathcal{W}^d \cdot h_k))}
\end{gather}
where $\mathcal{W}^d$ denotes the weights for estimating attention coefficient, $\Gamma$ is a single layer feed-forward network. The model uses masked softmax to obtain the normalized attention coefficients $\mu^d_{i, j}$.

In this work, we adopt a three-layer GAT to propagate node features on the consistency graph. The input is a node feature matrix $\textbf{Z} \in \mathbb{R}^{N \times C}$ given by a pre-trained ELMo \cite{peters2018deep}. After three layers of GATs, the node features are mapped to $D$ dimensions, which are the same with visual embeddings.

\begin{table}
\renewcommand\tabcolsep{4pt}
\newcommand{\beforefulloffset}{\hspace{-0.1cm}}
\newcommand{\beforeseenoffset}{\hspace{-0.6cm}}
\newcommand{\beforeunseenoffset}{\hspace{-1.1cm}}
\caption{HOI Detection results on HICO-DET dataset under zero-shot settings. \textit{UC}, \textit{UO} and \textit{UA} denote unseen action-object combination, unseen object and unseen action scenarios respectively.}
\label{tab:3}
\begin{tabularx}{\linewidth}{p{2.15cm}<{\raggedright}|p{0.7cm}<{\centering}|p{1.6cm}<{\centering}p{1.6cm}<{\centering}p{1.6cm}<{\centering}}
\toprule
\textbf{Method}&\textbf{Type}&\beforefulloffset\textbf{Full}&\beforeseenoffset\textbf{Seen}&\beforeunseenoffset\textbf{Unseen}\\
\midrule
Shen \textit{et al.} \cite{shen2018scaling}&\multirow{3}{*}{UC}&\beforefulloffset6.26&\beforeseenoffset-&\beforeunseenoffset5.62\\
Bansal \textit{et al.} \cite{bansal2019detecting}&&\beforefulloffset12.45$\pm$0.16&\beforeseenoffset12.74$\pm$0.34&\beforeunseenoffset11.31$\pm$1.03\\
\textbf{ConsNet} (ours)&&\beforefulloffset\textbf{19.81}$\pm$0.32&\beforeseenoffset\textbf{20.51}$\pm$0.62&\beforeunseenoffset\textbf{16.99}$\pm$1.67\\
\midrule
Bansal \textit{et al.} \cite{bansal2019detecting}&\multirow{2}{*}{UO}&\beforefulloffset13.84&\beforeseenoffset14.36&\beforeunseenoffset11.22\\
\textbf{ConsNet} (ours)&&\beforefulloffset\textbf{20.71}&\beforeseenoffset\textbf{20.99}&\beforeunseenoffset\textbf{19.27}\\
\midrule
\textbf{ConsNet} (ours)&UA&\beforefulloffset\textbf{19.04}&\beforeseenoffset\textbf{20.02}&\beforeunseenoffset\textbf{14.12}\\
\bottomrule
\end{tabularx}
\end{table}

\begin{figure*}
\centering
\includegraphics[width=\textwidth]{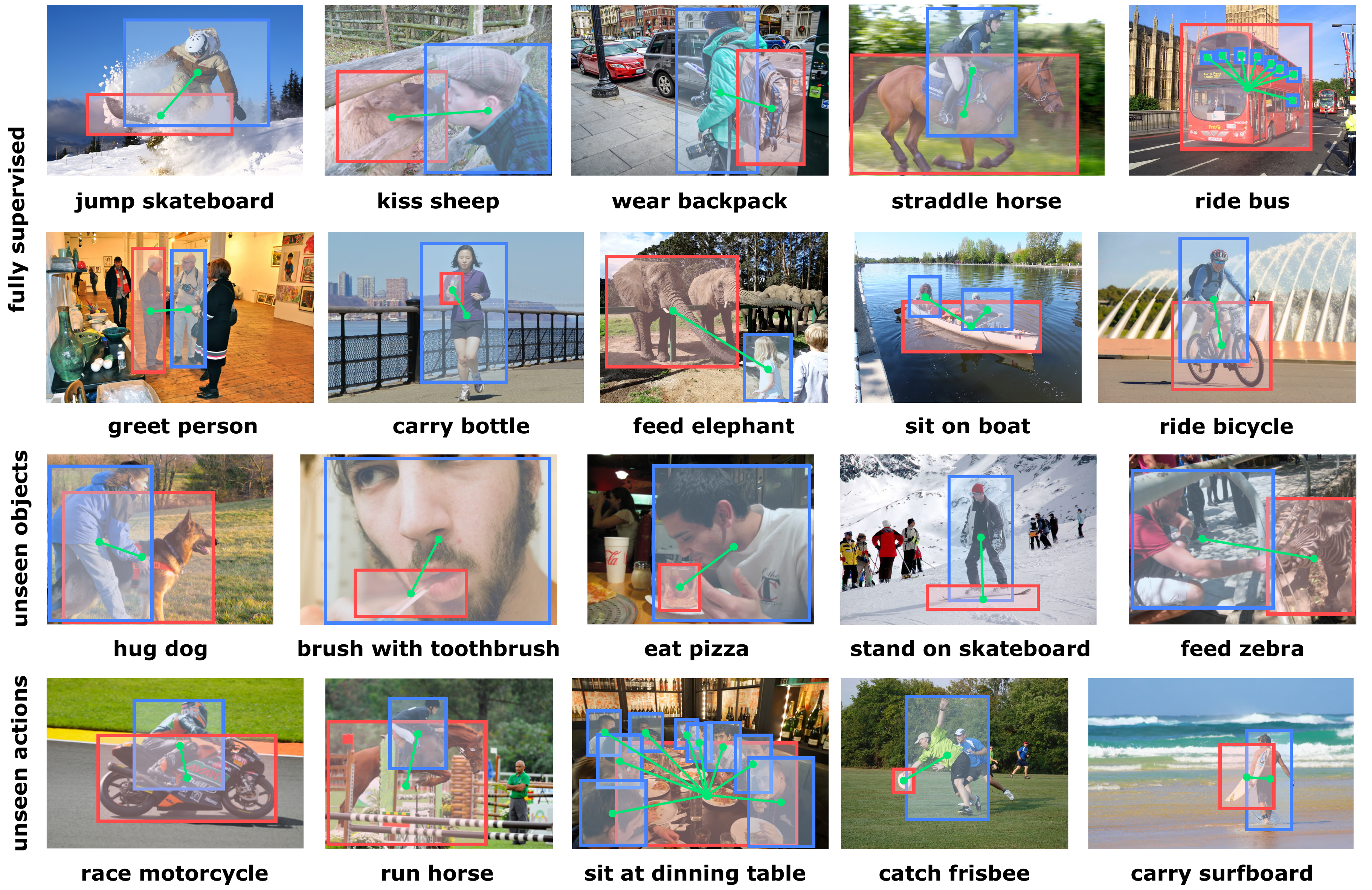}
\caption{Qualitative results on HICO-DET dataset. Our model has the ability to detect seen HOIs, HOIs with unseen objects and HOIs with unseen actions. Note that none of the previous models can detect HOIs with unseen actions.}
\Description{Qualitative results on HICO-DET dataset.}
\label{fig6}
\end{figure*}

\subsection{Model Learning}

During training, visual embedding network learns to map visual features of human-object pairs into visual-semantic joint embedding space, while semantic embedding network learns to generate semantic embeddings of HOI categories. When testing, the semantic embeddings can be pre-computed and be used as templates of HOI categories. Since all the proposed components are differentiable, the whole model can be trained in an end-to-end manner. The overall objective of training is to minimize the distance among visual embeddings and semantic embeddings. We learn the parameters of the whole model by supervising $r_{h, o}$ and $\varphi_{h, o}$ with the following binary cross-entropy losses:
\begin{gather}
\mathcal{L}_i = -(u \cdot log(\varphi_{h, o}) + (1 - u) \cdot log(1 - \varphi_{h, o}))\\
\mathcal{L}_c = -\frac{1}{C} \sum^C_{k = 1} (y_k \cdot log(r^k_{h, o}) + (1 - y_k) \cdot log(1 - r^k_{h, o}))
\end{gather}
where $u$ denotes interactiveness label and $y_k$ indicates HOI label. The interactiveness loss $\mathcal{L}_i$ and classification loss $\mathcal{L}_c$ are jointly optimized using their weighted sum by
\begin{gather}
\mathcal{L} = \mathcal{L}_i + \eta \cdot \mathcal{L}_c
\end{gather}
where $\eta$ is a scale factor balancing the loss weights. Note that we optimize the classification loss only with positive samples and the interactiveness loss with both positive and negative samples.

\section{Experiments}

In this section, we evaluate the proposed method on the challenging V-COCO \cite{gupta2015visual} and HICO-DET \cite{chao2018learning} datasets. We first evaluate our method under the fully-supervised settings on both of the datasets, following by zero-shot settings on HICO-DET dataset. The zero-shot settings includes three scenarios, i.e. unseen action-object combination, unseen object, and unseen action. An extensive ablation study is also reported after the evaluations.

\subsection{Datasets and Evaluation Metrics}

V-COCO is a subset of MS-COCO 2014 dataset \cite{lin2014microsoft}, it has 2,533 images for training, 2,867 images for validation and 4,946 images for testing. Each person is annotated with binary labels of 26 action categories. HICO-DET is another large-scale HOI detection dataset that extends annotations of HICO \cite{chao2015hico} from image-level to instance-level. The \verb|trainval| split has 38,118 images while the \verb|test| split has 9,658 images. It contains 117 action classes for 80 object classes, resulting in 600 HOI categories.

We follow the standard evaluation metric introduced by Chao \textit{et al.} \cite{chao2018learning} that uses mean average precision (mAP) to measure the detection performance. An HOI detection is considered as a true positive when both the bounding boxes of the human and object have intersection over union (IoU) with a ground truth greater than 0.5, and the predicted HOI label is correct.

\subsection{Implementation Details}

We adopt Faster R-CNN \cite{ren2015faster} with ResNet-50-FPN as the object detector. The same backbone and neck are also used for feature extraction. We train the object detector on MS-COCO 2017 dataset using MMDetection \cite{chen2019mmdetection} and then freeze its weights. When training the HOI classifier, we use all the detections with confidence greater than 0.1 and make use of both ground truths and the detected candidate pairs. When testing, we only consider up to 10 humans with confidence greater than 0.5 and up to 20 objects with confidence greater than 0.1 per image to reduce computational cost.

We add batch normalization \cite{ioffe2015batch} and ReLU nonlinearity \cite{glorot2011deep} after all hidden layers. The classification losses of different samples are weighted to prevent overfitting. Each training mini-batch contains 64 samples with the ratio of positive and negative samples $1 : 3$. For all experiments, we use Stochastic Gradient Descent (SGD) optimizer with initial learning rate 0.01, momentum 0.9, and weight decay 0.0001. The linear warm-up policy starting from 0.001 learning rate for 500 iterations is adopted. All the models are trained for 5 epochs using cosine annealing learning rate schedule.

\subsection{Fully-Supervised HOI Detection}

We first evaluate our model under fully-supervised settings. For both datasets, we train the model on \verb|trainval| split and evaluate it on \verb|test| split. The comparisons on V-COCO and HICO-DET datasets are shown in Table~\ref{tab:1} and Table~\ref{tab:2}. Our method outperforms the previous best models on each subset. Note that for HICO-DET dataset, the object detectors in Bansal \textit{et al.} \cite{bansal2019detecting} and PPDM \cite{liao2020ppdm} are trained on MS-COCO and finetuned on HICO-DET, which may provide more potential true positives and largely reduce false positives. To be directly comparable, we also report the performance of our model with a finetuned detector called \textit{ConsNet-F}, indicating that our method still achieves higher performance.

\begin{table}
\renewcommand\tabcolsep{4pt}
\caption{Ablation study results on HICO-DET dataset under fully-supervised settings. $\varnothing$ means predicting HOI labels using visual embedding network directly.}
\label{tab:4}
\begin{tabularx}{\linewidth}{p{0.95cm}<{\centering}|p{1.55cm}<{\centering}|p{1cm}<{\centering}|p{1cm}<{\centering}p{1cm}<{\centering}p{1cm}<{\centering}p{1cm}<{\centering}}
\toprule
\textbf{Type}&\textbf{Embedder}&\textbf{Depth}&\textbf{Full}&\textbf{Rare}&\multicolumn{2}{l}{\hspace{-0.15cm}\textbf{Non-Rare}}\\
\midrule
$\varnothing$&-&-&18.90&10.57&21.40\\
MLP&ELMo&3&19.01&11.82&21.15\\
SGC&ELMo&3&19.63&14.85&21.05\\
GCN&ELMo&3&20.15&15.12&21.66\\
SAGE&ELMo&3&20.07&15.05&21.58\\
\midrule
GAT&ELMo&2&21.16&16.82&22.46\\
\textbf{GAT}&\textbf{ELMo}&\textbf{3}&\textbf{22.15}&\textbf{17.55}&\textbf{23.52}\\
GAT&ELMo&4&21.12&16.35&22.54\\
\midrule
GAT&Word2Vec&3&20.59&15.94&21.98\\
GAT&GloVe&3&20.63&15.66&22.12\\
GAT&FastText&3&20.58&15.68&22.04\\
\bottomrule
\end{tabularx}
\end{table}

\subsection{Zero-Shot HOI Detection}

Shen \textit{et al.} \cite{shen2018scaling} first introduced the concept of zero-shot HOI detection that detects HOIs with unseen action-object combinations, where the actions and objects are seen in other HOIs. Bansal \textit{et al.} \cite{bansal2019detecting} proposed to detecting HOIs with unseen objects. We now extend the task further and introduce the scenario of detecting HOIs with unseen actions, which means the model should have the ability to analogize semantic representations of new actions based on similar actions or interactions, which is much more challenging than the two scenarios above. Below we report the performance comparisons under these scenarios on HICO-DET dataset.

\subsubsection{Unseen Combination Scenario}

The first three rows in Table~\ref{tab:3} shows the comparison of our method with others under unseen combination scenario. We use the same 5 sets of 120 unseen classes as Bansal \textit{et al.} and report the means of the results. The comparison shows that our approach does much better on detecting unseen HOIs with seen actions and objects.

\subsubsection{Unseen Object Scenario}

Line 4\;\textasciitilde\;5 in Table~\ref{tab:3} presents the performance comparison under unseen object scenario. Our model marginally outperforms the previous best method on unseen classes while having similar performance on seen classes, indicating that our method can better generalize to unseen objects.

\subsubsection{Unseen Action Scenario}

In this scenario, we randomly select 22 actions, define them as \verb|unseen| and remove all the training samples containing these actions. The full list of unseen actions will be publicly available. We then train the model on the remaining samples and evaluate on the full \verb|test| split. The last row in Table~\ref{tab:3} reports the performance of our approach on detecting HOIs with unseen actions. The results show that our model has the ability to detect HOIs even if the action is previously unseen, which is challenging because transferring the knowledge of actions is much harder than objects. Moreover, our approach can even do slightly better than some early methods under fully-supervised settings.

\subsection{Qualitative Results}

Figure~\ref{fig6} shows qualitative results of both fully-supervised and zero-shot HOI detection using our method. Even if our model has never seen the objects or actions before, the semantic embedding network can still benefit from seen HOIs and generate semantic representations of unseen HOIs.

\subsection{Ablation Study}

In order to analyze the significance of the proposed knowledge-aware strategy for generating semantic representations, we evaluate the models with different styles of semantic embedding networks and types of language models. All the experiments are performed on HICO-DET dataset under fully-supervised settings and the results are shown in Table~\ref{tab:4}.

Compared with not using semantic embedding network and simply using an MLP, HOI detection results on rare classes are largely improved with the use of GNNs. This is because the aggregation functions of GNNs can help transfer knowledge from non-rare classes to rare ones. The comparison also shows that with learnable attention coefficients, GATs are more flexible than other GNNs for generating semantic embeddings. Besides, the number of GAT layers matters. Deeper GATs can bring more learnable parameters, while it may cause the over-smoothing problem \cite{li2019deepgcns}, leading to a performance drop. Performances are also considerably improved by changing word embeddings from Word2Vec \cite{mikolov2013efficient}, GloVe \cite{pennington2014glove}, or FastText \cite{joulin2016bag} to ELMo \cite{peters2018deep}. The probable reason is that ELMo can better capture information at trigram level since the triplet is considered jointly as a whole.

\section{Conclusion}

In this work, we propose an end-to-end trainable framework for knowledge-aware human-object interaction detection by incorporating a consistency graph and exploiting GATs to propagate knowledge among nodes. Leveraging such a graph structure and message passing strategy, the model can capture and transfer knowledge about HOIs at different granularities and better generate semantic representations for rare or previously unseen HOIs.

\begin{acks}

This research is supported in part by \grantsponsor{0}{Key-Area Research and Development Program of Guangdong Province, China}{} with Grant \grantnum{0}{2019B010155002}, \grantsponsor{1}{National Natural Science Foundation of China}{} Grant \grantnum{1}{91538203}, \grantsponsor{2}{US NSF}{} Grant \grantnum{2}{1405594}, and start-up funds from University at Buffalo.

\end{acks}

\bibliographystyle{ACM-Reference-Format}
\bibliography{main}

\end{document}